# Discriminative Learning via Semidefinite Probabilistic Models

**Koby Crammer**
Department of Computer and Information Science
University of Pennsylvania
Philadelphia, PA 19104

**Amir Globerson**
CSAIL
Massachusetts Institute of Technology
Cambridge, MA 02139

## Abstract

Discriminative linear models are a popular tool in machine learning. These can be generally divided into two types: linear classifiers, such as support vector machines (SVMs), which are well studied and provide state-of-the-art results, and probabilistic models such as logistic regression. One shortcoming of SVMs is that their output (known as the "margin") is not calibrated, so that it is difficult to incorporate such models as components of larger systems. This problem is solved in the probabilistic approach. We combine these two approaches above by constructing a model which is both linear in the model parameters and probabilistic, thus allowing maximum margin training with calibrated outputs. Our model assumes that classes correspond to linear subspaces (rather than to half spaces), a view which is closely related to concepts in quantum detection theory. The corresponding optimization problems are semidefinite programs which can be solved efficiently. We illustrate the performance of our algorithm on real world datasets, and show that it outperforms second-order kernel methods.

## 1 Introduction

Support Vector Machines (SVM) [12] are commonly regarded as the state of the art in supervised learning. One of their key advantages is their maximization of the margin, a property which also guarantees certain generalization bounds [1]. A different class of supervised learning algorithms is the one based on probabilistic models, such as logistic regression [9]. Such models are parameterizations of the class conditional distributions, which are trained to maximize the likelihood of the observed data. One advantage of probabilistic models is that the probabilities they generate may be used as a calibrated measure of certainty about class prediction. Such a measure may be used in systems which incorporate classifiers as modules, and is generally useful in balancing different types of errors. The confidence measure in SVMs is commonly taken to be the margin of an example, which is a geometric quantity and is not naturally calibrated or even bounded.

While there have been previous attempts on assigning probabilistic outputs to SVMs [11], they have been based on transforming the margin into the $[0, 1]$ range, and not on a complete probabilistic model.

Another integral property of SVMs is of course the half-space structure of classes (in the binary case). An equivalent statement is that SVMs assume there is a transformation of inputs into the real line such that positive and negative points correspond to different classes. Moreover, by using kernels, linear separation need only be assumed for a nonlinear transformation of the variables. However, geometric intuition is often lost as a result of the kernel transformation, and the resulting separators are not easily interpretable.

In this work, we present a different view of class separation, which incorporates both the concepts of margin maximization *and* probabilistic modeling. Our approach assumes that classes correspond to orthogonal linear subspaces in feature space. This assumption is reasonable in many domains where the existence or absence of a feature is the key predictor of its class identity, rather than its exact value or its relation to values of other features. For example, in document classification there may be subsets of words (or linear combinations of word counts) whose appearance indicates the document topic. In image classification, a set of pixels may be indicative of image content regardless of their exact intensity ratios. An alternative statement of the problem is that there exists a linear transformation of feature space such that a unique subset of coordinates is active in each class.

In order to measure the degree to which a given input point *belongs* to a given subspace we use a projection operator which measures what fraction of the point's norm lies in a subspace. Such projection operators correspond to matrices with eigenvalues in the discrete set $\{0, 1\}$. We relax this assumption to the $[0, 1]$ range, which makes the model tractable. It also turns out that the output of the projection operators have a natural interpretation as probabilities, and these probabilities are *linear* functions of the model parameters (the projection matrices).

Because the model is both a linear and a probabilistic model, we can efficiently implement both methods that rely on margin maximization, and those that maximize probability related measures such as log likelihood or optimal Bayes errors. All these problems are convex and two of them are Semidefinite Programs (SDP) [16] for which efficient algorithms exist.

Our model is closely related to ideas in quantum detection and estimation, where semidefinite matrices are used to generate probabilities. A simple view is that the class conditional models are represented by SDP matrices with a unit trace and the detectors are represented using PSD matrices.

We compare the performance of our method to the closely related second order kernels SVM, and show that it achieves improved performance on a handwritten digit classification task, while providing meaningful probabilistic output.

## 2 The Probabilistic Model

Consider a classification task where $\mathbf{x} \in \mathbb{R}^d$ are the feature vectors, and classes are $y \in \{1, \ldots, k\}$. Denote the classification rule by $f(\mathbf{x}) = y$. We assume that classes reside in linear subspaces $S_y$ (i.e., $\mathbf{x} \in S_y \Leftrightarrow f(\mathbf{x}) = y$) such that $S_i \cap S_j = \{0\}$ ($\forall i \neq j$), $S_i$ and $S_j$ are orthogonal spaces, and the spaces $S_y$ span the entire space: $S_1 \oplus S_2 \ldots \oplus S_k = \mathbb{R}^d$. This corresponds to the assumption that there exists a linear transformation in feature space such that a subset of coordinates is active for a given class, and these coordinates are mutually exclusive.

The projection operator on the space $S_y$ is a matrix $A_y$ such that $A_y$ is idempotent (thus, for every $\mathbf{x} \in S_y$ we have $A_y \mathbf{x} = \mathbf{x}$) and symmetric ($A_y^2 = A_y, A_y = A_y^T$). This implies that if $\mathbf{x} \in S_y$ then $\|A_y \mathbf{x}\|^2 = \|\mathbf{x}\|^2$, and if $\mathbf{x} \notin S_y$ then $\|A_y \mathbf{x}\|^2 \leq \|\mathbf{x}\|^2$. The above suggests that $\|A_y \mathbf{x}\|^2$ may be taken as a measure of the degree to which $\mathbf{x}$ *belongs* to class $y$.

Now, note that $\|A_y \mathbf{x}\|^2 = \mathbf{x}^T A_y \mathbf{x}$ so that this measure is in fact a quadratic function of $\mathbf{x}$, and importantly is *linear* in $A_y$.

Since we are interested in the multiclass setting, it is natural to define $\|A_y \mathbf{x}\|^2$ as the probability of class $y$ given the point $\mathbf{x}$:

$$p(y|\mathbf{x}) = \frac{1}{\mathbf{x}^T (\sum_y A_y) \mathbf{x}} \mathbf{x}^T A_y \mathbf{x} . \qquad (1)$$

This implies the probability is invariant to the norm of $\mathbf{x}$ and we can thus always normalize $\mathbf{x}$ such that $\|\mathbf{x}\|^2 = 1$.

The normalization factor in Eq. (1) makes the probability a nonlinear function of $A_y$. However, our assumption on the structure of the classes in fact implies that $\sum_y A_y = I$. To see this, denote the orthogonal basis of $S_y$ by $V_y$. The assumption about the structure of $S_y$ implies that $\bigcup_y V_y = \{\mathbf{v}_1, \ldots, \mathbf{v}_d\}$ yields an orthogonal basis of the entire space. Denote by $V$ the matrix whose columns are $\{\mathbf{v}_i\}_{i=1}^n$. Then $\sum_y A_y = \sum_i \mathbf{v}_i \mathbf{v}_i^T = VV^T = I$ by the assumption of orthogonality and the fact that $V_y$ is orthogonal to $V_{y'}$ for $y \neq y'$.

Since $\sum_y A_y = I$ the probabilistic model of Eq. (1) reduces to

$$p(y|\mathbf{x}) = \mathbf{x}^T A_y \mathbf{x} . \qquad (2)$$

Optimization over the set of idempotent matrices is an integer optimization problem, which seems to be hard to solve. We therefore relax this assumption, and only constrain $A_y$ to be positive semidefinite, and to satisfy $\sum_y A_y = I$. These two constraints imply that the eigenvalues of $A_y$ lie between zero and one. Since idempotent matrices are characterized by eigenvalues in $\lambda \in \{0, 1\}$, we can interpret our relaxation as relaxing this eigenvalues constraint by the constraint $\lambda \in [0, 1]$ (see e.g. [14]).

## 3 The Learning Problem

We now turn to the problem of learning a classifier using the probabilistic model defined in Eq. (2). Given a labeled sample $(\mathbf{x}_i, y_i)$, $i = 1, \ldots, n$, we seek a set of parameters $A_y$ which result in a *good* classifier. Below we present two approaches to this problem. The first is related to margin based methods, and the second to likelihood based ones.

### 3.1 Margin based approaches

A desired property in a classifier is that the probability it assigns to the correct class is higher that those assigned to incorrect classes. In other words, we wish to maximize the margin between the correct probability $p(y_i|\mathbf{x}_i)$ and the incorrect ones $p(z|\mathbf{x}_i)$ where $z \neq y_i$. Define the margin of a point $\mathbf{x}_i$ by,

$$m_i = p(y_i|\mathbf{x}_i) - \max_{z \neq y_i} p(z|\mathbf{x}_i) \ .$$

Then, as in other margin based classifiers, we wish to maximize the minimum margin. In the separable case (i.e., there exists a classifier such that all margins on the training set are positive), the margin maximization problem is given by the following semidefinite program:

$$\begin{array}{ll} \max & \eta \\ \text{s.t} & p(y_i|\mathbf{x}_i) - p(z|\mathbf{x}_i) \geq \eta \quad \forall i, \quad z \neq y_i \\ & \sum_y A_y = I \\ & A_y \succeq 0 \end{array}$$

If the data is not separable, we add a slack variable $\xi_i$ for each sample point

$$\begin{array}{ll} \max & \eta - \beta \sum_i \xi_i \\ \text{s.t} & p(y_i|\mathbf{x}_i) - p(z|\mathbf{x}_i) \geq \eta - \xi_i \quad \forall i, \quad z \neq y_i \\ & \sum_y A_y = I \\ & A_y \succeq 0, \xi_i \geq 0 \end{array} \quad (3)$$

where $\beta \geq 0$ is a tradeoff parameter.

We call this method the MaxMargin approach since it seeks a maximum margin model.

### 3.2 Likelihood based approaches

Since our model is a parametric family of distributions, one way to optimize it is via standard maximum likelihood. This yields the following optimization problem

$$\begin{array}{ll} \max & \sum_i \log p(y_i|\mathbf{x}_i) \\ \text{s.t} & \sum_y A_y = I \\ & A_y \succeq 0 \end{array} \quad (4)$$

Note that since $p(y_i|\mathbf{x}_i)$ is a linear function of the parameters, its log is concave, and the optimization problem is thus concave, although it is not a standard SDP, since the objective is nonlinear. We do not study this approach further in this manuscript, since we prefer to focus on problems for which standard solvers exist.

A related approach is obtained if we consider the measure of success of the predictor to be the probability it assigns to the correct class. This view implies that we should perform the following maximization

$$\begin{array}{ll} \max & \sum_i p(y_i|\mathbf{x}_i) \\ \text{s.t} & \sum_y A_y = I \\ & A_y \succeq 0 \end{array} \quad (5)$$

This optimization is very similar to the maximum likelihood one, but without the log function. The objective can also be viewed as the optimal Bayes loss in prediction given that the true distribution is $p(y|x)$. We therefore denote this optimization by Bayes.

Note that for logistic models such as Conditional Random Field (CRF) [9], this problem is not convex since CRF probabilities are not convex functions of the parameters. Interestingly, this problem may be solved analytically for the binary case as we now show.

Denote the two matrices by $A_1$ and $A_2 = I - A_1$. The constraints imply that the eigenvalues of $A_1$ are between zero and one. The objective function then becomes,

$$\sum_{i:y_i=1} \text{tr}(A_1 \mathbf{x}_i \mathbf{x}_i^T) + \sum_{i:y_i=2} \text{tr}((I - A_1)\mathbf{x}_i \mathbf{x}_i^T) \ .$$

Omitting the constant term which does not affect the solution we get,

$$\text{tr}\left(A_1 \left(\sum_{i:y_i=1} \mathbf{x}_i \mathbf{x}_i^T - \sum_{i:y_i=2} \mathbf{x}_i \mathbf{x}_i^T\right)\right) \ .$$

Let $\mathbf{v}_i$ and $\lambda_i$ be the eigenvectors and the eigenvalues of the constant matrix,

$$\sum_{i:y_i=1} \mathbf{x}_i \mathbf{x}_i^T - \sum_{i:y_i=2} \mathbf{x}_i \mathbf{x}_i^T \ .$$

Since the objective is linear in the matrix $A_1$ we get that $A_1$ has the same eigenvectors $\mathbf{v}_i$. Let $d_i$ be the eigenvalues of $A_1$. We have that the objective function is given by

$$\text{tr}\left(A_1 \left(\sum_{i:y_i=1} \mathbf{x}_i \mathbf{x}_i^T - \sum_{i:y_i=2} \mathbf{x}_i \mathbf{x}_i^T\right)\right) = \sum_i \lambda_i d_i \ .$$

Therefore, to maximize the objective function one should set $d_i = \text{sgn}(\lambda_i)$, where we define $\text{sgn}(0) = 0.5$. To conclude, we showed that the solution of Eq. (5) for two classes can be obtained by computing the difference between the (normalized) covariances matrices per class, and assigning each of the eigenvectors to one of the matrices $A_y$ in accordance with the sign of the corresponding eigenvalue. A similar algorithm was proposed in the context of quantum detection theory where more information is assumed. See the book of Helstrom [8] for more details.

## 4 Convex bounds on the zero-one loss

A common approach to choosing an optimal classifier is to find the one which minimizes a convex upper bound on the zero one loss. In conditional log-linear models [9], the function $-\log_2 p(y|\mathbf{x})$ is such a convex upper bound (convex in the model parameters). In

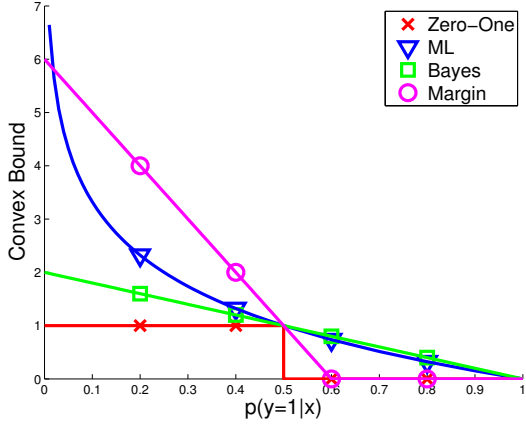

Figure 1: Convex upper bounds on the zero one loss in the binary case. Curves are shown for the case where $y = 1$ is the correct class. For the margin bound, a value of $\eta = 0.2$ is used.

Support Vector Machines [12] the hinge loss serves as a bound.

The zero-one loss is given by

$$l_{zo}(\mathbf{x}, y, p) = \Theta[p(y|x) - 0.5] \qquad (6)$$

To illustrate the bounds in our models, we focus on the binary class case. Figure 1 shows the bounds discussed below and their relation to the zero-one loss.

The simplest linear upper bound on the zero-one loss is (see Figure 1)

$$l_{\text{Bayes}}(\mathbf{x}, y, p) = 2(1 - p(y|\mathbf{x})) \qquad (7)$$

As its name suggests, it is minimized by the Bayes optimization problem in Eq. (5).

The ML problem in Eq. (4) corresponds to minimizing the bound (see Figure 1)

$$l_{\text{ML}}(\mathbf{x}, y, p) = -\log_2(p(y|\mathbf{x})) \qquad (8)$$

The interpretation of the maximum margin formulation is slightly more complex. Consider the function

$$l_{\text{Marg}}(\mathbf{x}, y, p, \eta) = \max\{0, 1 + \frac{1}{\eta} - \frac{2}{\eta} p(y|x)\} \qquad (9)$$

The function $l_{\text{Marg}}$ is also an upper bound on the zero-one loss, as can be seen in Figure 1 and is similar to the hinge loss, with the exception that the former is parameterized by $\eta$.

The objective in Eq. (3) in the binary case can then be written as a sum of elements

$$\eta(1 - \beta l_{\text{Marg}}(\mathbf{x}_i, y_i, p, \eta)) \qquad (10)$$

For $\beta = 1$ the factor in the parenthesis may be interpreted as a lower bound on the probability of correct classification. Thus the max-margin method may be viewed as optimizing a (multiplicative) tradeoff between correct classification and margin maximization. Different values of $\beta$ reflect the weight that should be attributed to classification rate compared to margin.

## 5 Duality

As in the case of SVM, convex duality may be used to gain important insights into the problem. We obtain the convex dual of Eq. (3) by introducing two sets of dual parameters. The first (corresponding to the normalization constraint) is $\lambda$, a matrix of size $d \times d$. The second (corresponding to the margin constraints) is $q_{yi}$ ($y = 1, \ldots, k$ and $i = 1, \ldots, n$) where we force $q_{y_i i} = 0$.

Standard duality transformation yields the dual semidefinite program

$$\begin{aligned}
\min \quad & -\text{tr}(\lambda) \\
\text{s.t.} \quad & \sum_{i,z} q_{zi} = 1 \\
& \sum_z q_{zi} \leq \beta \quad \forall i = 1, \ldots, n \\
& \sum_{i:y_i \neq y} q_{yi} \mathbf{x}_i \mathbf{x}_i^T - \sum_{i:y_i = y} (\sum_z q_{zi}) \mathbf{x}_i \mathbf{x}_i^T \succeq \lambda
\end{aligned} \qquad (11)$$

where the last constraint is true for all $y$.

We can use the dual to further interpret the meaning of the $\beta$ parameter. Assume that $\beta = 1/(\nu n)$ where $n$ is the number of examples and $\nu \in [0, 1]$. We say that the $i^{th}$ example is *not* a support vector if the solution of Eq. (11) satisfies $q_{zi} = 0$ for all $z$. Intuitively, an example which is not a support vector does not change the solution of the optimization problem and thus can be *omitted*, without affecting the solution. Note that this definition is somewhat weaker than the standard definition in support vector machines, since we do not have a representer theorem that links the primal and dual solutions. We also say that the i$th$ example is a margin error if $\xi_i > 0$. The following lemma links the value of $\nu$ to both margin errors and the number of support vectors and is analogous the $\nu$-property in [13] (proposition 12).

**Lemma 5.1 :** *Let $(\eta, \xi_i, A_y)$ be the solution of the primal optimization problem and let $(q_{yi}, \lambda)$ be the solution of the dual. Then,*

1. *$\nu$ is an upper bound on the fraction of margin errors.*

2. *$\nu$ is a lower bound on the fraction of support vectors.*

**Proof:** At most a fraction of $\nu$ examples can satisfy $\sum_z q_{zi} = \beta = 1/(\nu n)$. This is because $\sum_{i,z} q_{zi} = 1$. But from KKT conditions we know that $\sum_z q_{zi} = \beta$ if $\xi_i > 0$. Hence the first part of the lemma. Any support vector can contribute at most a mass of $1/(\nu m)$ to the sum $\sum_{i,z} q_{zi} = 1$. Thus, there are at least $\nu n$ examples which are support vectors. $\square$

## 6 Implementation Issues

The semidefinite programs discussed above can be solved using existing solvers such as CSDP [3]. This package was used in the experiments discussed below. However, for large $n$ or $d$ this approach becomes impractical. An alternative approach, which yielded similar results, is to use a projected sub-gradient algorithm [2]. The projected sub-gradient algorithm takes small steps along the sub-gradient of the objective, followed by Euclidean projection on the set of constraints.

To see how it may be applied, note that Eq. (3) may be written as

$$\begin{aligned} \max \quad & \eta - \beta \sum_i [\eta - p(y_i|\mathbf{x}_i) + \max_{z \neq y_i} p(z|\mathbf{x}_i)]_+ \\ s.t \quad & \sum_y A_y = I \\ & A_y \succeq 0 \end{aligned}$$

Thus the objective is a non-differentiable function, and the only constraints are positivity of $A_y$ and the normalization constraints. It is straightforward to obtain the sub-gradient of the objective. We now turn to the Euclidean projection part of the algorithm.

Here the set of constraints is given by

$$\begin{aligned} S_{norm} &= \left\{ A_y : \sum_y A_y = I \right\} \\ S_{pos} &= \left\{ A_y : A_y \succeq 0 \right\} \\ S &= S_{norm} \cap S_{pos} \end{aligned}$$

Define the Euclidean projection of the parameters $A_y$ on $S$ by

$$\{A_y^p\} = \arg \min_{\hat{A}_y \in S} \sum_y \|A_y - \hat{A}_y\|^2 \qquad (12)$$

In the binary class case, this projection can be found analytically. Define the matrix $C = (A_1 - A_2 + I)/2$, and denote by $\mathbf{v}_i, \lambda_i$ its eigenvectors and eigenvalues. Then it can be shown that the projection is given by

$$A_1^p = \sum_i \min(1, \max(0, \lambda_i)) \mathbf{v}_i \mathbf{v}_i^T \quad , \quad A_2^p = I - A_1$$

In the multiclass case, there does not seem to be an analytic solution. However, since projection on each of the sets $S_{norm}, S_{pos}$ is straightforward, one can use Dykstra's alternating projection algorithm [6] to obtain the Euclidean projection on $S$.

## 7 Relation to $2^{nd}$ order kernel methods

Our probabilistic model is closely related to SVM with second order kernels. To see this, note that $\mathbf{x}^T A_y \mathbf{x} = \text{tr}(A_y \mathbf{x}\mathbf{x}^T)$ which may be interpreted as a dot product between the elements of $A_y$ and $\mathbf{x}\mathbf{x}^T$. This is precisely the form of the predictor obtained for SVM with a second order kernel. There are however several key differences between our approach and the SVM one. The first is that the outputs in our case are automatically normalized probabilities, whereas the SVM need not even be positive. The second is that the bound on the zero-one loss used in our learning algorithm is significantly different from that used in SVM.

Clearly, the class of models we learn are a subclass of those available to second order SVMs, due to the constraints on the matrices $A_y$. To gain more insight into the constraints, consider the case where $A_y$ is constrained to be diagonal. The resulting classification rule will be based on the dot product between $\text{diag}(A_y)$ and the element-wise square of $\mathbf{x}$. Since the diagonal elements will then be constrained to be in the range $[0, 1]$, this case corresponds to a linear SVM on the squared $\mathbf{x}$ with box constraints on the weight vectors. This creates an interesting link between our method and linear separators with positive weights such as the Winnow algorithm [10].

The relation between the log-likelihood formulation (Eq. (4)) and logistic-regression is not direct as the relation between the margin formulation (Eq. (3)) and SVMs. This is because in our model, probabilities are linear in the parameters, while for logistic regression they are obtained through exponentiation of linear terms.

In the experimental evaluations below, we compared our method to second order SVMs, and found that the former achieved better performance. We elaborate on possible reasons for this result in what follows.

## 8 Quantum Mechanics

The formulation we proposed, and especially the Likelihood based approach, are related to analogous detection problems in the quantum mechanics literature. We begin with some definitions. A *density operator* $\rho$ is a positive semi-definite matrix, with a unit trace, $\text{tr}(\rho) = 1$. We can think of the density operator as defining a distribution over the eigenvec-tors of the operators, with a weight proportional to the corresponding eigenvalue. Specifically, denote by $\rho = \sum_i \nu_i \mathbf{v}_i \mathbf{v}_i^T$ where $\|\mathbf{v}_i\|^2 = 1$. Then,

$$\Pr[\mathbf{v}_i] = \nu_i = \mathbf{v}_i^T \rho \mathbf{v}_i = \mathrm{tr}(\rho \mathbf{v}_i \mathbf{v}_i^T) \ .$$

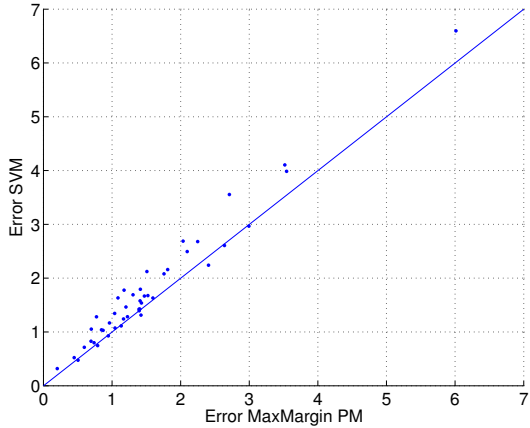

Figure 2: Test error (in percentages) of the MaxMargin algorithm (x-axis) vs. test error (in percentages) of SVM (y-axis) for all the 45 label-pairs of the USPS dataset. A point above the line $y = x$ indicates better performance for the MaxMargin algorithm.

We can also use the density operator to define a probability measure over every normalized vector $\mathbf{x}$ using the same algebraic form and have, $\Pr[\mathbf{x}] = \mathrm{tr}(\rho \mathbf{x} \mathbf{x}^T)$.

We now turn our attention to the problem of quantum hypothesis testing [8]. Assume that there are given $k$ density-operators $\rho_i$ for $i = 1 \ldots k$. Our goal is to find a set of $k$ operators $\Pi_i$ which we shall call *detection operators*. These operators are positive semi-definite whose sum is the identity, $\sum_i \Pi_i = I$. These detection operators are used to define the conditional detection probabilities,

$$\Pr[\text{state } j \,|\, \text{state i}] = \mathrm{tr}(\rho_i \Pi_j)$$

that the detectors choose the j*th* state when the i*th* state is true. Let us denote by $\zeta_j$ the prior probability of being in j*th* state. Then, the average detection error is given by,

$$\sum_j^k \sum_i^k \zeta_j (1 - \delta_{i,j}) \mathrm{tr}(\rho_j \Pi_i) \ ,$$

where $\delta_{i,j} = 1$ if $i = j$ and $\delta_{i,j} = 0$ if $i \neq j$. The goal of the system designer is to find a set of detection operators $\Pi_i$ that will minimize the average error. Eldar [7] proposed a few formulations of the problem as semi definite-programs. Note that the above problem is similar to our Likelihood formulation in Eq. (4).

In the machine-learning formulation given in the current work, we do not assume to be given either the prior probabilities $\zeta_i$ nor the density operators $\rho_i$, but only a finite sample from both. Specifically, we assume to have only pairs of a vector $\mathbf{x}$ and a label $y$. Where the label $y$ was drawn in accordance to the prior $\zeta_i$ and the vectors $\mathbf{x}$ in accordance to the class conditional probabilities $\rho_i$.

## 9 Related Work

A few attempts were made to combine large margin classifiers with probabilistic outputs. The most notable example is the work of Platt [11]. This work suggests using a sigmoid on the outputs of the support vector machine, and provides ways to calibrate the parameters of this sigmoid.

An alternative approach was discussed by Cesa-Bianchi et al in [4, 5]. They suggested to force the output of a linear classifier to the [0, 1] range (and thus have a probabilistic interpretation) by assuming that both the input vector $\mathbf{x}$ *and* the weight vector $\mathbf{v}$ lie in a ball of radius one. Thus the value of the inner product between the weight vector and the input vector is always in the range $[-1, 1]$, which is mapped linearly into the range $[0, 1]$. Note that there is no simple and direct extension of this approach into multi-class problems.

Several directions which relate machine learning and quantum mechanics were proposed recently. Warmuth [17] presents a generalization of the Bayes rule to the case when the prior is a density matrix. Wolf [18] provides interesting relations between spectral clustering and other algorithms based on Euclidean distance, and the Born rule. Our likelihood based approach is related to some of the many detection algorithms presented by Eldar [7]. Note that unlike Eldar, we do not assume direct knowledge of a probabilistic model (prior probabilities or density operators) but only a finite sample from it.

## 10 Experimental Evaluation

To illustrate how our method extracts subspaces from data, we first apply it to a simple two dimensional XOR problem shown in Figure 3. The resulting PSD matrices $A_y$ turn out to be projection matrices (i.e. eigenvalues in $\{0, 1\}$) although their eigenvalues could be non-integers in principle. Furthermore, we have $p(y_i | x_i) = 1$ for all sample points.

We next evaluated our algorithm using the USPS handwritten digits dataset. The training set contains $7,291$ training examples and the test set has $2,007$ ex-amples. Originally, each instance represents an image of a size $16 \times 16$ of a digit. There are ten possible digits. Since our current implementation (CSDP) is still limited in the data size it can handle, we reduced the dimensionality of the data by replacing each four adjacent pixels with their mean, resulting in image of size $8 \times 8$. Thus, the dimensionality was reduced from 256 to 64. We enumerated over all 45 pairs of digits and repeated the following process 10 times. For each pair we randomly chose 300 examples which were associated with one of the two digits of the current pair. The remaining training examples associated with this pair were used as a validation set. The test set was the standard USPS test set (restricted to the relevant two digits).

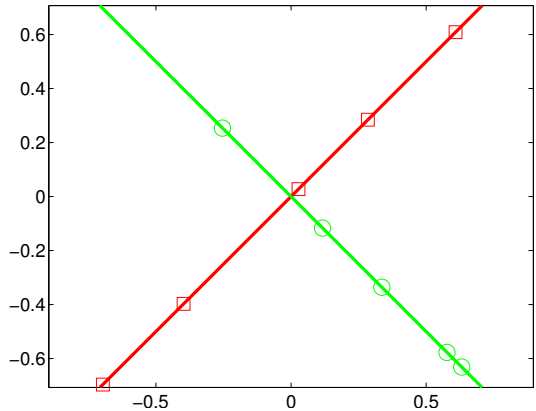

Figure 3: An example of subspace learning in two dimensions. The classes in this case are the two one dimensional subspaces (i.e., lines) corresponding to the vectors $\mathbf{v}_1 = [1, 1], \mathbf{v}_2 = [-1, 1]$. The sample points (5 points per class) are drawn randomly from these lines. Applying our max-margin algorithm with $\beta = 0.1$ to this sample results in matrices $A_1, A_2$ with spectra $[0, 1], [1, 0]$ respectively. The lines corresponding to the dominant directions in each $A_y$ are shown in the figure.

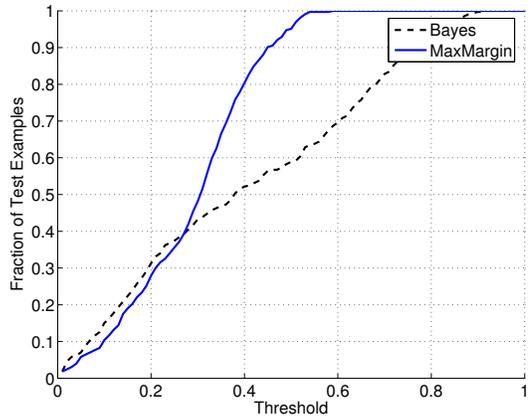

Figure 4: Fraction of examples in test set which the difference in probability $|p(3|\mathbf{x}) - p(5|\mathbf{x})|$ is below a threshold set by a value in the x-axis.

We trained three algorithms: support vector machines (SVMs), our maximum-margin formulation in Eq. (3) (denoted by MaxMargin) and our optimal Bayes formulation in Eq. (5) (denoted by Bayes). For SVMs we used 9 values for the regularization parameter $\beta$ and for the MaxMargin method we tried 6 values for the regularization parameter. We trained each of the algorithms using all the values of the parameter and picked the one model which achieved minimal error over the validation set. We then used this model to compute the error over the test set. We averaged the results over the 10 repeats.

Figure 2 summarizes the results for both SVMs and the MaxMargin approach. Each point corresponds to one of the 45 binary classification problems. A point above the line $y = x$ corresponds to a pair where MaxMargin performs better than SVMs, and vise-versa. Clearly, MaxMargin outperforms SVMs, as most of the points are above the line $y = x$. We computed a similar plot for the Bayes algorithm which turned out to be worse than both MaxMargin and SVMs.

To better understand the performance of Bayes and MaxMargin we focus our attention on one of the 45 binary problems. Specifically, we chose the hard task of discriminating between the digits *three* and *five*. This is the hardest task for SVMs. We picked one of the partitions of the data into training-set and validation-set and computed the absolute difference in probability for each of the test examples: $|p(5|\mathbf{x}) - p(3|\mathbf{x})|$. We then enumerated over several possible threshold values of this difference, and recorded the fraction of test examples for which this difference is higher than the value of the threshold. The results are summarized in Figure 4 for the MaxMargin and Bayes algorithms. As one can observe from the figure, for the MaxMargin algorithm all of the examples have a difference in probability that is less than 0.6. While for the Bayes algorithm the difference of the probabilities is even as high as 0.8. This result can be explained by the following observation: The goal of the MaxMargin algorithm is to maximize the number of correct predictions. For this task, there is no need to have a high-difference in probabilities, only high-enough difference (of about 0.5). On the other hand, the Bayes algorithm optimizes the expected error when drawing a label using the probability model $p(y|\mathbf{x})$. It thus tries to push the probabilities apart from each other, even at the cost of making some prediction error. Indeed, this is the case here, since Bayes generally yields worse generalization error than MaxMargin. However, its probabilities seem to *better calibrated*, suggesting that Bayes should in some cases be the preferred algorithmic choice.

## 11 Discussion

We presented algorithms for learning subspaces using probabilistic models. This resulted in semidefinite optimization, and allowed both max-margin and likelihood objectives.

Note that although our presentation referred to the case of orthogonal subspaces, a much wider class of subspaces are separable under our classification rule (intuitively, subspaces such that the angle between them is above 45 degrees in the binary case).

The empirical results presented above show that our method compares favorably with second order SVM. Since our model is effectively a subclass of the latter, it is not immediately clear why this should be the case. There are two differences between our method and SVMs which could shed light on these results. The first is that since we optimize over a constrained parameter set for the weights, generalization error variance is reduced, albeit at the cost of possibly increased bias. It will be very interesting to obtain theoretic results in this respect. While it does not seem like the VC dimension of our class is smaller than the corresponding SVM, there still may be theoretical guarantees which result from our constraints (positive semidefiniteness and normalization) on parameter space. Another possible explanation for the empirical results is the difference in the objective function, and related convex bounds on the zero-one loss. While SVM uses a hinge loss to bound the zero-one loss, our method effectively uses the bounds discussed in Section 3.2. One difference between these two bounds, is that the hinge loss heavily penalizes points with negative margin, whereas in our case this penalty is upper bounded.

An interesting extension of our method is to model local interactions via semidefinite matrices. This would correspond to Taskar's extension of SVM to the multi-label case [15], and would hopefully share the probabilistic interpretation of Conditional Random Fields [9].

**Acknowledgments** The authors thank the Rothschild Foundation - Yad Hanadiv for their generous support.